# SSSDET: SIMPLE SHORT AND SHALLOW NETWORK FOR RESOURCE EFFICIENT VEHICLE DETECTION IN AERIAL SCENES

*Murari Mandal, Manal Shah, Prashant Meena, Santosh Kumar Vipparthi*

Vision Intelligence Lab, Malaviya National Institute of Technology Jaipur

## ABSTRACT

Detection of small-sized targets is of paramount importance in many aerial vision-based applications. The commonly deployed low cost unmanned aerial vehicles (UAVs) for aerial scene analysis are highly resource constrained in nature. In this paper we propose a simple short and shallow network (SSSDet) to robustly detect and classify small-sized vehicles in aerial scenes. The proposed SSSDet is up to 4× faster, requires 4.4× less FLOPs, has 30× less parameters, requires 31× less memory space and provides better accuracy in comparison to existing state-of-the-art detectors. Thus, it is more suitable for hardware implementation in real-time applications. We also created a new airborne image dataset (ABD) by annotating 1396 new objects in 79 aerial images for our experiments. The effectiveness of the proposed method is validated on the existing VEDAI, DLR-3K, DOTA and Combined dataset. The SSSDet outperforms state-of-the-art detectors in term of accuracy, speed, compute and memory efficiency.

## 1. INTRODUCTION

The evolution of unmanned aerial vehicles (UAVs) from being a niche technology used for military applications to becoming a capable and affordable tool for consumer and industrial applications has opened a new frontier of computer vision, i.e., aerial vison, which requires analysis and interpretation of aerial images and videos. Aerial vision-based data is being generated in abundance in both consumer and industrial market space. These aerial images and video data facilitate numerous applications such as aerial surveillance, search and rescue, event recognition, urban and rural scene understanding. An essential low-level task for the abovementioned applications is to detect the objects, particularly vehicles, on the ground. Vehicle detection in aerial images is a challenging task due to the variable sizes of the vehicles (small, medium and large), high/low density of vehicles and complex background in the cameras field of view. Moreover, the aerial scenes in urban setup usually comprises of a varieties of object types leading to excessive interclass object similarities. These similarities between the target and nontarget objects makes it very difficult to distinguish between the vehicles and nonvehicle objects in aerial images. Furthermore, the recent advancements in the development of affordable UAVs have created a strong need for object-detection algorithms that can operate on resource constrained environment.

Vehicle detection in aerial images is a well-studied problem and the literature for the same can be divided into two categories: designed and learned feature-based methods. The methods in the first category extract the visual features using hand-crafted feature descriptors (i.e. Haar-features, SIFT, LBP, HOG, Gabor filters, etc.) [1-3]. These features are then used to detect and localize the objects with a classifier or cascade of classifiers. In [4], the authors proposed a sliding window mechanism to apply the filters at different positions and scales of an image. Uijlings et al. [5] introduced a selective search approach to generate the possible locations for an object and perform sampling based on the image structure. This approach has been widely used to generate candidate region proposals for further processing through SVM or neural network-based classifiers.

The methods in the second category have utilized convolution neural networks to learn features from an image for object detection. These methods can be categorized into two-stage and single-stage frameworks. Recent approaches [6-10] for aerial images have primarily used the two-stage architectures (fast/faster R-CNN [11]) based frameworks to detect vehicles in aerial scenes. The faster R-CNN consists of a region proposal network (RPN) and object detection network, leading to significant computational cost. Redmon et al. [12] proposed a unified one-stage model named YOLO to perform object detection and classification. Further, they proposed YOLOv2 [13] to improve the performance by introducing batch normalization, high resolution classifier, anchor boxes, dimension clustering and multiscale-training. More recently, YOLOv3 [14] and RetinaNet [15] were proposed to detect the smaller objects as well. However, these techniques are more suitable for images captured from canonical views. Also, these methods consist of a large number of parameters and require high memory space.

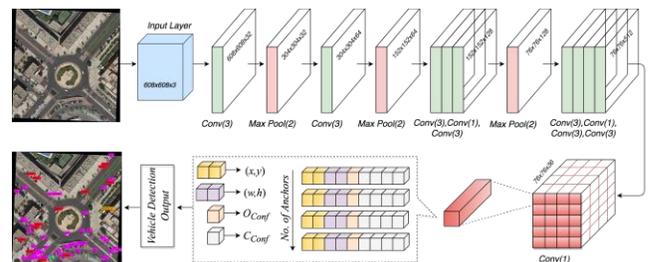

**Fig. 1.** The proposed SSSDet architecture is shown for object detection and classification in 4-class DOTA dataset. The final layer features are composed of 5776 tensors of size 1×1×36. Each tensor contains the bounding box coordinates (*x, y, w, h*), object confidence ($O_{Conf}$) and class confidence ($C_{Conf}$) for every anchor box.

In addition, to effectively deal with the challenges of rotation variations and appearance ambiguity in geospatial scenes, various rotation-invariant detectors [3, 16-18] have been proposed in the literature. Diao et al. [19] proposed saliency-based object detection using deep belief networks. Nie et al. [8] used multi-task models to combine the semantic labelling and detection information for more accurate detection results.

In this paper, we proposed a lightweight network for vehicle detection in aerial scenes. The SSSDet is a simple short and shallow convolutional network optimized for fast inference and high accuracy. It is designed according to the intuition and ideas that have appeared in the recent literature, some of which we incorporate to develop a robust and resource efficient vehicle detector in aerial images.

The proposed SSSDet preserves the small object features by using fewer down-sampling and convolutional layers to learn salient object characteristics. This also ensures fewer number of trainable parameters and model size as compared to the more popular object detectors like YOLOv3 and RetinaNet. To detect the densely populated objects, we generate enlarged feature maps in the final layer of the network. Moreover, the input layer is enlarged to maintain higher object-to-pixels ratios.

The SSSDet outperforms the state-of-the-art object detection approaches in both accuracy and resource (computation and memory space) efficiency. Thus, it can be highly suitable for resource constraint systems such as UAVs, offering excellent accuracy and efficiency.

## 2. PROPOSED SSSDet

The proposed SSSDet performs object localization and classification in a single step. The underlying convolutional network generates feature maps for an input image, which corresponds to a fixed-sized tensor (depending on the number of classes and anchors [13]). The final feature map is trained to learn bounding-box coordinate offsets, object and class confidence scores for every individual anchor.

The SSSDet architecture is presented in Fig. 1. It consists of ten convolutional layers and three down sampling (max pooling) layers. The max pooling is performed with non-overlapping 2×2 windows. We use only two convolutional layers for initial high-resolution maps of 608×608 and 304×304. Afterwards, two lightweight blocks of convolution (3×3,1×1,3×3) operations are applied to learns the high-level features for small-sized vehicles. Each block consists of a 1×1 projection that reduces the dimensionality. The final feature map is generated by applying a 3×3 convolution with depth corresponding to the

**Table 1.** Summarization of the evaluated datasets

| Dataset | #Images | #Objects | #Object per class |
|---|---|---|---|
| VEDAI | 1248 | 3773 | car: 1393, truck: 307, pickup: 955, tct: 190, cc: 397, bt: 171, mc: 4, bus: 3, van: 101, other: 204, large: 48 |
| DLR-3K | 262 | 8401 | car: 8210, hv: 191 |
| DOTA | 1558 | 55235 | car: 24516, hv: 11307, pln: 4733, bt: 14679 |
| ABD | 79 | 1396 | car: 1353, hv: 11, bt: 32 |
| Complete | 3099 | 68579 | car: 36510, hv: 12406, pln: 4781, bt: 14882 |

*tct: tractor, cc: camping car, mc: motorcycle, hv: heavy vehicle, pln: plane, bt:boat

number of anchors and object classes. In order to enhance the delineation of small-sized objects, we generated bigger feature maps in the final layer of the network.

We place Batch Normalization and Leaky-ReLU between all convolutions. We do not use bias in any of the projection in order to reduce the number of parameters and overall memory requirement. This choice didn't have any adverse impact on the performance of SSSDet. Due to the shallowness of our network, the SSSDet consists of only 1.99 million parameters which is 30 times less than YOLOv3.

### 2.1. Design Choices

Even with increased input layer size, the objects in aerial images are usually quite small with lower object-to-pixels ratios. Too much down sampling of the feature maps may lead to vanishing of small object features. To understand this, let's take an image of size 1024×1024 and a vehicle object of size 40×40 inside this image. If the image is resized (to 608×608) and further down sampled 32 times after several convolution and max pooling layers. This will result in negligible feature representation (around 1 or 2 scalar values) for the small-sized vehicles in the output feature maps, which is insufficient to accurately detect that object. This in turn, may cause localization of the objects in the original image very difficult. Since, the final layer predicts the bounding boxes for the presence of an object, we need various features of an object to be adequately represented (spatial dimension of the final convolutional feature map). Therefore, in our proposed work, we have generated convolutional feature maps of size 76×76. Moreover, through a set of experiments on parameter sensitivity analysis on computational performance and model accuracy, we configured the input layer size as 608×608.

### 2.2. Implementation Details

We train the SSSDet using darknet [24, 25] framework over 1 Titan Xp GPU. The network is optimized using stochastic gradient descent (SGD) with minibatch size of 4. The weight decay and momentum parameters are set to 0.0005 and 0.9. The threshold for non-maximum suppression is set to 0.6. The loss is calculated by taking the sum of square error from the final layer of the network as given in [12, 13]. For each aerial scene dataset, we train the SSSDet from scratch without using any pretrained model weights to initialize the network. The network predicts the object probability, class probability and bounding box coordinate offsets for every anchor at the final layer as shown in Fig. 1.

**Table 2.** Comparative detection performance in terms of mean average precision (mAP) of the proposed SSSDet and existing state-of-the-art approaches

| Method | VEDAI | DLR | DOTA | Complete |
|---|---|---|---|---|
| YOLOv2_416 | 9.08 | 9.61 | 33.36 | 28.86 |
| YOLOv2_608 | 25.12 | 26.81 | 47.45 | 48.04 |
| Faster R-CNN | 34.82 | 20.04 | 42.29 | 38.02 |
| YOLOv3_416 | 32.07 | 52.11 | 74.46 | 70.35 |
| YOLOv3_608 | 38.98 | 54.49 | 76.60 | 75.21 |
| RetinaNet | 43.47 | 54.77 | 73.77 | 71.28 |
| YOLOv3-tiny_416 | 11.10 | 26.42 | 47.88 | 46.73 |
| YOLOv3-tiny_608 | 31.73 | 39.74 | 65.89 | 59.17 |
| **SSSDet** | **45.97** | **58.25** | **79.52** | **77.22** |

## 3. EXPERIMENTAL RESULTS AND DISCUSSIONS

In this section, we evaluate the performance of our model on multiple aerial image datasets. The performance of SSSDet is analyzed in terms of mAP, compute and space complexity.

### 3.1. Datasets

We evaluate the performance of our model on VEDAI [20], DLR-3K [1], DOTA [22] and the Complete dataset (VEDAI, DLR-3K, DOTA and ADB). For VEDAI, the annotations provided in [21] were used. For the remaining datasets DLR-3K, DOTA and ADB, we have manually annotated all the images and generated horizontal bounding boxes. The 20 images in DLR-3K were each divided into 16 parts to generate 320 images. We annotated the objects in DLR-3K into two categories: car, heavy vehicle. We also annotated the DOTA objects with four categories (car, heavy vehicle, plane, boat) as opposed to 15 categories given by the authors. We collected 79 new aerial images from online sources and generated a new dataset named airborne dataset (ABD) by annotating 1396 objects for our experiments. Furthermore, a large dataset was augmented by combining VEDAI, DLR-3K, DOTA and ABD datasets. This would enable a more comprehensive performance analysis of the proposed and existing object detectors in aerial scenes. The summary description of all the datasets is given in Table 1. We plan to release all these annotations in future for public use.

### 3.2. Experimental Setup

The VEDAI, DOTA and Complete dataset were divided in train and test set with a ratio of ~ [90:10]. Similarly, we used ~ [80:20] division for model training and performance evaluation in DLR-3K dataset. All the existing and the proposed approaches were evaluated with the same setup. The SSSDet network was trained for ~30k-50k iterations on VEDAI, DOTA and Complete dataset. The initial learning rate is set to 0.001 which is reduced by a factor of 10 after 20k iterations. Similarly, the model was trained for ~20k iterations over DLR-3K. The initial learning rate of 0.001 was reduced by a factor of 10 after 10k iterations. SSSDet generates four bounding boxes corresponding to each grid cell and selects the bounding box with the highest IoU with respect to a given threshold. The RetinaNet [25] and Faster R-CNN [11] detectors were trained over each aerial dataset using pretrained weights (over ImageNet) ResNet-50 and ResNet-101 respectively. We have used the detection framework given in [23, 24] to train and evaluate the SSSDet, YOLOv2 and YOLOv3.

### 3.3. Results and Analysis

The comparative performance of the proposed SSSDet and other state-of-the-art approaches in terms of mAP for vehicle detection in the VEDAI, DLR-3K, DOTA and the complete dataset is given in Table 2. The mAP corresponds to the average of the maximum precisions at different recall values. Moreover, the precision-recall graphs at different IoU thresholds for SSSDet, YOLOv2 and YOLOv3 for each dataset is shown in Fig. 2. From Fig. 2 and Table 2, it is evident that the SSSDet outperforms YOLOv2, YOLOv3, Faster RCNN and RetinaNet detectors in all the datasets. More specifically, it achieves 6.99%, 3.76%, 2.92% and 2.01% mAP improvement over YOLOv3_608.

### 3.4. Analysis for Real Time Applications

In Table 3, we report a comparison of number of floating-point operations, parameters and memory space required by different models. SSSDet efficiency is evident, as it requires 4.4× less FLOPs, has 30× less parameters, requires 31× less memory space and provides better accuracy in comparison to YOLOv3_608. It also has 4× less parameters and model size as compared to the lightweight version of YOLOv3. However, as shown in Table 2, the accuracy of YOLOv3-tiny is significantly lower than SSSDet.

We also report inference speed of SSSDet, YOLOv3 and YOLOv3-tiny on a CPU system with Intel i5 2.5 GHz processor and 4GB RAM. The proposed SSSDet is 3.75× faster than YOLOv3 as shown in Table 3. Although, YOLOv3-tiny has better inference speed, it is significantly less accurate as compared to proposed SSSDet. Therefore, from Table 2 and Table 3, it is evident, that the proposed

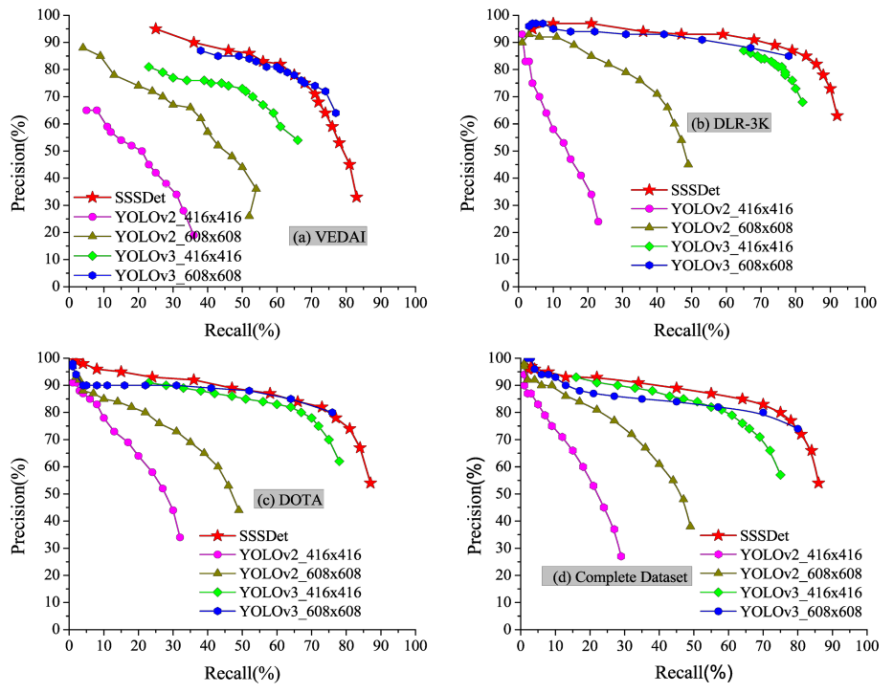

**Fig. 2.** Precision-recall graph of the proposed SSSDet and existing state-of-the-art object detectors over (a) VEDAI, (b) DLR-3K, (c) DOTA and (d) Complete dataset.

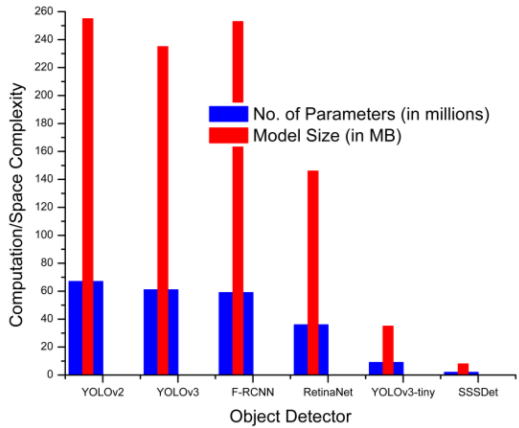

**Fig. 3.** Computation and space complexity comparison of the proposed SSSDet with the existing state-of-the-art object detectors.

SSSDet outperforms existing state-of-the-art techniques in terms of overall performance.

## 4. CONCLUSIONS

We have proposed a resource efficient vehicle detection technique named SSSDet for aerial scene. Our main objective was to make efficient use of scarce resources available on mobile platforms such as UAVs as compared to fully fledged GPU systems. SSSDet preserves the salient features of the small-sized objects by generating enlarged feature maps through a simple short and shallow network. The proposed SSSDet achieve significant gain in resource (computation and memory) efficiency while exceeding the existing state-of-the-art object detectors in terms of accuracy. It consists of a significantly smaller number of FLOPs and is much faster than YOLOv3. In addition, we created a new dataset ABD by collecting 79 new aerial images (annotated 1396 objects) from open sources. We demonstrated the efficacy of the SSSDet by conducting experiments on four challenging datasets VEDAI, DLR-3K, DOTA and Complete dataset. The SSSDet outperforms the existing state-of-the-art approaches in terms of mAP, computation (no. of parameters) and space (model size) complexity.

**Table 3.** Computation and space complexity comparison of the SSSDet with YOLOv3 and YOLOv3-tiny with input layer size=608×608×3. *The FPS is computed over a CPU system*

| Method | BFLOPS | Param | Model Size | FPS |
|---|---|---|---|---|
| YOLOv3 | 139.52 | 61.6 M | 235 MB | 0.04 |
| YOLOv3-tiny | **5.42** | 8.8 M | 35 MB | **0.40** |
| **SSSDet** | 32.29 | **1.9 M** | **8 MB** | 0.15 |

## 5. ACKNOWLEDGEMENTS

This work was supported by the Science and Engineering Research Board (under the Department of Science and Technology, Govt. of India) project #SERB/F/9507/2017. The authors would like to thank the members of Vision Intelligence Lab and Kautilya Bhardwaj for their valuable support. We are also thankful to NVIDIA for providing TITAN Xp GPU grant.